\documentclass[10pt]{article}

\usepackage{microtype}
\usepackage{graphicx}
\usepackage{booktabs}
\usepackage[colorlinks=true,linkcolor=blue,citecolor=blue,urlcolor=blue]{hyperref}
\usepackage[utf8]{inputenc}
\usepackage[T1]{fontenc}

\usepackage[letterpaper,margin=1in]{geometry}

\usepackage{amsmath}
\usepackage{amssymb}
\usepackage{mathtools}
\usepackage{bm}

\graphicspath{{figures/}{figures/experiments/}}

\makeatletter
\def\input@path{{./}{setup/}{includes/}{figures/}{figures/experiments/}}
\makeatother

\usepackage{etoolbox}
\usepackage{xifthen}
\usepackage{subcaption}
\usepackage{textcomp}
\usepackage{yhmath}
\usepackage[ruled,vlined]{algorithm2e}
\usepackage{cancel}
\usepackage{xcolor}
\usepackage{pifont}
\usepackage{textcase}
\usepackage{siunitx}
\usepackage{array}
\usepackage{multirow}
\usepackage{gensymb}
\usepackage{tabularx}
\usepackage{extarrows}
\usepackage{colortbl}
\usepackage{pgfplots}
\usepackage{pgfplotstable}
\usepackage{adjustbox}
\usepackage{doi}
\usepackage{xurl}
\providecommand{\address}[2][]{}  % article class compatibility (custom_commands expects it)
\usepackage{setup/custom_commands}

\pgfplotsset{compat=newest}
\usetikzlibrary{bayesnet,fadings,patterns,shadows.blur,calc,shapes,3d,backgrounds,fit,arrows.meta,positioning,shapes.geometric}
\allowdisplaybreaks
\definecolor{darkgreen}{rgb}{0.0, 0.5, 0.0}


\usepackage[numbers,sort&compress]{natbib}
\begin{document}

\noindent\rule{\textwidth}{1pt}
\begin{center}
    {\fontsize{14pt}{17pt}\selectfont\bfseries
        Joint Parameter and State-Space Bayesian Optimization:\\
        Using Process Expertise to Accelerate Manufacturing Optimization
    }
\end{center}
\noindent\rule{\textwidth}{1pt}
\begin{center}
    {\large
        Saksham Kiroirwal$^{1}$, Julius Pfrommer$^{1}$, J\"urgen Beyerer$^{1,2}$\par
        \footnotesize
        $^{1}$Cognitive Industrial Systems, Fraunhofer IOSB, Karlsruhe, Germany\par
        $^{2}$Karlsruhe Institute of Technology (KIT), Karlsruhe, Germany\par
        \texttt{\{saksham.kiroirwal,julius.pfrommer,juergen.beyerer\}@iosb.fraunhofer.de}
    }
\end{center}
\begin{abstract}
    \noindent
    Bayesian optimization (BO) is a powerful method for optimizing black-box manufacturing processes, but its performance is often limited when dealing with high-dimensional multi-stage systems, where we can observe intermediate outputs. Standard BO models the process as a black box and ignores the intermediate observations and the underlying process structure. Partially Observable Gaussian Process Networks (POGPN) model the process as a Directed Acyclic Graph (DAG). However, using intermediate observations is challenging when the observations are high-dimensional state-space time series. Process-expert knowledge can be used to extract low-dimensional latent features from the high-dimensional state-space data. We propose POGPN-JPSS, a framework that combines POGPN with Joint Parameter and State-Space (JPSS) modeling to use intermediate extracted information. We demonstrate the effectiveness of POGPN-JPSS on a challenging, high-dimensional simulation of a multi-stage bioethanol production process. Our results show that POGPN-JPSS significantly outperforms state-of-the-art methods by achieving the desired performance threshold twice as fast and with greater reliability. The fast optimization directly translates to substantial savings in time and resources. This highlights the importance of combining expert knowledge with structured probabilistic models for rapid process maturation.
\end{abstract}
\noindent\textbf{Keywords:} Bayesian Optimization, Partially Observable Gaussian Process Network, Manufacturing Optimization, Joint Parameter and State-Space Optimization.

\section{Introduction}\label{sec:intro}
The goal of a manufacturing process is to increase the throughput and reduce expenses and operational bottlenecks~\cite{goldratt2006goal}. A common bottleneck is unoptimized or immature processes, which lead to billions of dollars in annual costs, high resource consumption, or waste or lost production time~\cite{ids2025realcost}. In many industrial settings, the primary objective is to reach a performance threshold—not necessarily perfection—in the shortest possible time. Achieving the threshold faster also minimizes resource consumption and opportunity costs.

\begin{figure}
      \centering
      \tikzset{every picture/.style={line width=0.7pt}} % Set default line width
\tikzset{plate caption/.append style={left=2pt of #1.south east, yshift=0.2cm}}
\begin{tikzpicture}[
            block/.style={rectangle, draw, rounded corners, minimum height=4em, minimum width=4.5em, thick},
            normal_arrow/.style={-Stealth},
            obs/.style={circle, draw, minimum size=6.5mm, align=center, inner sep=0.7pt},
            sum/.style={circle, draw, minimum size=4mm, align=center, inner sep=0pt},
            every node/.style={font=\normalsize},
            every picture/.style={line width=1pt}
      ]

      % Nodes
      \node[block, align=center] (process1) {};
      \node[block, align=center, right=of process1, xshift=-0.2cm] (process2) {};
      \node[block, align=center, right=of process2, xshift=-0.2cm] (process3) {};
      % Top-right in-box state labels
      \node[anchor=north east] at ($(process1.north east)+(-2pt,-2pt)$) {$\cusmatrix{\Obsstates}^{(1)}$};
      \node[anchor=north east] at ($(process2.north east)+(-2pt,-2pt)$) {$\cusmatrix{\Obsstates}^{(2)}$};
      \node[anchor=north east] at ($(process3.north east)+(-2pt,-2pt)$) {$\cusmatrix{\Obsstates}^{(3)}$};
      % Labels below boxes
      \node[above=2pt of process1, align=center] (process1_label) {Flask: $\process^{(1)}$};
      \node[above=2pt of process2, align=center] (process2_label) {Seed: $\process^{(2)}$};
      \node[above=2pt of process3, align=center] (process3_label) {Production: $\process^{(3)}$};

      \node[right=of process3, xshift=-0.5cm] (y3) {$\cusvector{\obsouts}^{(\objnode)}$};

      % Inputs and outputs
      \node[left=of process1, xshift=1.0cm, yshift=0.55cm] (input1) {$\cusvector{\inpvars}^{(1)}$};
      \node[right=of process1, xshift=-1.0cm, yshift=0.55cm] (input2) {$\cusvector{\inpvars}^{(2)}$};
      \node[right=of process2, xshift=-1.0cm, yshift=0.55cm] (input3) {$\cusvector{\inpvars}^{(3)}$};

      \node[draw=none, below=of process1, yshift=0.8cm] (f1_dummy) {};

      \draw[draw=darkgreen, normal_arrow] (input1) -- ++(0.0,-0.25) -- ($(process1.west) + (0,0.3)$);
      \draw[draw=darkgreen, normal_arrow] (input2) -- ++(0.0,-0.25) -- ($(process2.west) + (0,0.3)$);
      \draw[draw=darkgreen, normal_arrow] (input3) -- ++(0.0,-0.25) -- ($(process3.west) + (0,0.3)$);

      % Arrows between blocks and observations
      \draw[draw=red, normal_arrow] ($(process1.east) + (0,-0.25)$) -- node[below] (f1){$\cusvector{\trueouts}^{(1)}$} ($(process2.west) + (0,-0.25)$);
      \draw[draw=red, normal_arrow] ($(process2.east) + (0,-0.25)$) -- node[below] (f2){$\cusvector{\trueouts}^{(2)}$} ($(process3.west) + (0,-0.25)$);
      \draw[draw=blue, normal_arrow] (process3) -- (y3);
      % Time-series drawn relative to each block (normalized [0,1] coords)
      % Process 1
      \begin{scope}[
                  shift={(process1.south west)},
                  x={($(process1.south east)-(process1.south west)$)},
                  y={($(process1.north west)-(process1.south west)$)}
            ]
            \clip (0,0) rectangle (1,1);
            \draw[black!85, dashed] (0.10,0.15) -- (0.92,0.15);
            \draw[black!85, dashed] (0.10,0.15) -- (0.10,0.90);
            \draw[purple!85, thick, smooth] plot coordinates {
                        (0.10,0.75) (0.32,0.82) (0.56,0.60) (0.7,0.50) (0.85,0.30)
                  };
            \draw[orange!75, thick, smooth] plot coordinates {
                        (0.10,0.28) (0.38,0.38) (0.55,0.35) (0.7,0.38) (0.85,0.45)
                  };
      \end{scope}

      % Process 2
      \begin{scope}[
                  shift={(process2.south west)},
                  x={($(process2.south east)-(process2.south west)$)},
                  y={($(process2.north west)-(process2.south west)$)}
            ]
            \clip (0,0) rectangle (1,1);
            \draw[black!85, dashed] (0.10,0.15) -- (0.92,0.15);
            \draw[black!85, dashed] (0.10,0.15) -- (0.10,0.90);
            \draw[purple!85, thick, smooth] plot coordinates {
                        (0.10,0.25) (0.36,0.38) (0.62,0.52) (0.80,0.58) (0.85,0.62)
                  };
            \draw[orange!75, thick, smooth] plot coordinates {
                        (0.10,0.90) (0.20,0.50) (0.30,0.40) (0.50,0.23) (0.85,0.20)
                  };
      \end{scope}

      % Process 3
      \begin{scope}[
                  shift={(process3.south west)},
                  x={($(process3.south east)-(process3.south west)$)},
                  y={($(process3.north west)-(process3.south west)$)}
            ]
            \clip (0,0) rectangle (1,1);
            \draw[black!85, dashed] (0.10,0.15) -- (0.92,0.15);
            \draw[black!85, dashed] (0.10,0.15) -- (0.10,0.90);
            \draw[purple!85, thick, smooth] plot coordinates {
                        (0.1,0.20) (0.50,0.30) (0.80,0.65)
                  };
            \draw[orange!75, thick, smooth] plot coordinates {
                        (0.10,0.45) (0.20,0.40) (0.30,0.20) (0.50,0.23) (0.85,0.20)
                  };
      \end{scope}

      \plate [dashed, inner sep=0cm] {process} {(process1) (process2) (process3) (process1_label) (process2_label) (process3_label) (f1) (f2) (input1) (input2) (input3) (y3) (f1_dummy)} {Process: $\Process$};

\end{tikzpicture}
      \caption{A multi-stage bioethanol production process $\Process$ with three subprocesses: $\process^{(1)}$, $\process^{(2)}$, and $\process^{(3)}$, each with controllable inputs $\cusvector{\inpvars}=\left(\cusvector{\inpvars}^{(1)},\cusvector{\inpvars}^{(2)},\cusvector{\inpvars}^{(3)}\right)$. Using sensors, we are also able to observe the state-space $\cusmatrix{\Obsstates}^{(1)}$, $\cusmatrix{\Obsstates}^{(2)}$ and $\cusmatrix{\Obsstates}^{(3)}$ of the subprocesses respectively. The final process performance, $\cusvector{\obsouts}^{(\objnode)}$, is measured at the objective node $\objnode$.}
      \vspace{-0.75cm}
      \label{fig:multi_process}

\end{figure}
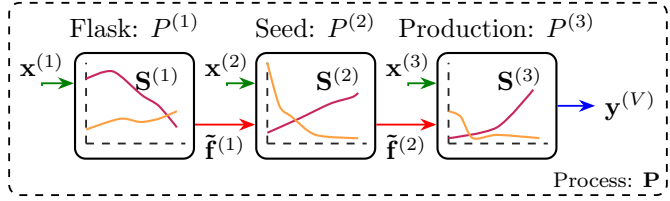

Figure~\ref{fig:multi_process} shows a multi-stage bioethanol production process as described by~\cite{hernandez2013seed}. We discuss the bioethanol production process in detail in Section~\ref{sec:problem_description}. Conventionally, such a system is modeled as a black-box $\Process$ that takes controllable inputs $\cusvector{\inpvars}$ and yields a product, whose performance is quantified by a metric $\cusvector{\obsouts}^{(\objnode)}$, such as the space-time yield (STY) of ethanol~\cite{sanchez2008bioethanol}. Bayesian optimization (BO) is a powerful black-box optimization technique widely used to maximize performance metrics in manufacturing~\cite{frazier2018tutorial, sano2020application, liang2021scalable, Shields2021Bayesian}.

However, modeling a process network as a simple black box can be suboptimal, as it ignores prior knowledge of its internal structure. As illustrated in Figure~\ref{fig:multi_process}, this process can be represented as a causal Directed Acyclic Graph (DAG) $\graph$, where nodes $\{\process^{(\numprocess)}\}_{\numprocess\in\numprocessset}$ correspond to individual subprocesses. We assume the DAG $\graph$ is topologically ordered, with the objective node $\process^{(\objnode)}$ being the final node. Each subprocess $\process^{(\numprocess)}$ receives controllable inputs $\cusvector{\inpvars}^{(\numprocess)}$ and the outputs $\cusvector{\trueouts}^{\nodeparent{\numprocess}}$ from its parent subprocesses $\nodeparent{\numprocess}$ to produce its own output $\cusvector{\trueouts}^{(\numprocess)}$. We are able to observe the outputs $\cusvector{\trueouts}^{(\numprocess)}$ as noisy observations $\cusvector{\obsouts}^{(\numprocess)}$ using sensors. However, we often observe high-dimensional state-space data $\cusmatrix{\Obsstates}^{(\numprocess)}$ within each subprocess as intermediate observations. For example, during bioethanol production, we can measure temperature and pH in the bioreactor over time.

Gaussian Process Networks (GPNs) introduced by ~\cite{astudillo2021bayesian,aglietti2020causal}, later extended by~\cite{kiroriwal2025partially} as Partially Observable GPN (POGPN), showed that BO can be improved by replacing the single-task GP surrogate model with a POGPN and incorporating prior knowledge of causal structure and intermediate observations $\cusvector{\intermediateobsouts}=\left(\cusvector{\obsouts}^{(\numprocess)}\right)_{\numprocess\in\numprocessset}$ of $\graph$. While powerful, both GPNs and POGPNs are not designed to handle high-dimensional state-space observations $\cusmatrix{\Obsstates}^{(\numprocess)}$, such as the time-series data common in modern manufacturing, which remains a key challenge.

Separately, the Joint Parameter and State-Space (JPSS) modeling proposed by~\cite{kiroriwal2024joint} showed how process expert knowledge can be used to extract low-dimensional latent features $\cusvector{\features}^{(\numprocess)}$ from the high-dimensional state-space $\cusmatrix{\Obsstates}^{(\numprocess)}$ to create a Gaussian Process Network (GPN). We present the following contributions:
\begin{itemize}
      \item Importance of process expert knowledge in BO and introduce POGPN-JPSS, a novel framework for optimizing complex processes with high-dimensional intermediate observations.
      \item Empirically show that POGPN-JPSS can rapidly mature the multi-stage bioethanol production process.
\end{itemize}
Section~\ref{sec:background} provides the background information on the Gaussian Processes and Bayesian Optimization. Section~\ref{sec:related_work} discusses the related work in detail. The described methods are used for the multi-stage bioethanol production process, described in Section~\ref{sec:problem_description}. Sections~\ref{sec:results} and~\ref{sec:benchmark_results} describe the experiments and results. Section~\ref{sec:discussion} discusses the results and provides the conclusion and outlook.
\section{Background}\label{sec:background}
We consider a manufacturing process $\Process$ composed of multiple subprocesses $\{\process^{(\numprocess)}\}_{\numprocess\in\numprocessset}$. When the process is evaluated at a given input configuration $\cusvector{\inpvars}_{\numobservation}=(\cusvector{\inpvars}^{(\numprocess)}_{\numobservation})_{\numprocess\in\numprocessset}$, we observe intermediate outputs $\cusvector{\intermediateobsouts}_{\numobservation}=(\cusvector{\obsouts}^{(\numprocess)}_{\numobservation})_{\numprocess\in\numprocessset}$ and a final objective output $\cusvector{\obsouts}^{(\objnode)}_{\numobservation}$. A dataset of $\Numobservation$ evaluations is denoted as $\mathcal{D}=\{\cusvector{\inpvars}_{\numobservation}, \cusvector{\intermediateobsouts}_{\numobservation},\cusvector{\obsouts}^{(\objnode)}_{\numobservation}\}^{\Numobservation}_{\numobservation=1}$.

\subsection{Stochastic Variational Inference for Gaussian Processes}\label{sec:SVGP}
A Single-Task Gaussian Process (STGP) is a Bayesian nonparametric model that defines a distribution over functions $\processfunction(\cdot)$~\cite{rasmussen2003gaussian}. For a given input $\cusvector{\inpvars}_{\numobservation}$, the function value $\processfunction^{(\objnode)}_{\numobservation}$ is assumed to be drawn from a Gaussian distribution $\probability(\processfunction^{(\objnode)}_{\numobservation}\vert\cusvector{\inpvars}_{\numobservation})=\mathcal{N}\big(\gpmean(\cusvector{\inpvars}_{\numobservation}), \gpkernel(\cusvector{\inpvars}_{\numobservation}, \cusvector{\inpvars}_{\numobservation}')\big)$, where $\gpmean(\cdot)$ is the mean function and $\gpkernel(\cdot,\cdot')$ is the covariance function. We also assume that the likelihood $\probability(\obsouts^{(\objnode)}_{\numobservation}\vert\processfunction^{(\objnode)}_{\numobservation})$ of observing $\cusvector{\obsouts}^{(\objnode)}_{\numobservation}$ is Gaussian. An STGP treats the entire process $\Process$ as a single black box, and hyperparameters are learned by maximizing the marginal log-likelihood (MLL)
\begin{equation}\label{eq:STGP_MLL_exact}
    \mathcal{L}_\text{GP} =
    \sum^{\Numobservation}_{\numobservation=1}\log\mathbb{E}_{\probability(\processfunction^{(\objnode)}_{\numobservation}\vert\cusvector{\inpvars}_{\numobservation})}\big[\probability(\obsouts^{(\objnode)}_{\numobservation}\vert\processfunction^{(\objnode)}_{\numobservation})\big].
\end{equation}
However, the cubic computational complexity of STGP inference makes it impractical for large datasets or hierarchical models~\cite{rasmussen2003gaussian, hensman2013gaussian}. The Stochastic Variational Gaussian Process (SVGP)~\cite{hensman2013gaussian} makes inference scalable by introducing a set of $\Numinducing$ inducing points $\cusvector{\inducingpoints}$, which are function values at chosen inducing locations $\cusmatrix{\Inducinglocations}$. These inducing points are summarized by a variational distribution $\variationalprob(\cusvector{\inducingpoints})=\mathcal{N}\big(\cusvector[bm]{\distmean}_{\cusvector{\inducingpoints}}, \cusmatrix[bm]{\distcov}_{\cusvector{\inducingpoints}}\big)$. For input $\cusvector{\inpvars}_{\numobservation}$, SVGP approximates the true GP posterior with a variational distribution $\variationalprob(\processfunction^{(\objnode)}_{\numobservation})=\int\variationalprob(\processfunction^{(\objnode)}_{\numobservation}\vert\cusvector{\inducingpoints})\variationalprob(\cusvector{\inducingpoints})\text{d}\cusvector{\inducingpoints}$. The GP hyperparameters and variational parameters ($\cusmatrix{\Inducinglocations}$, $\cusvector[bm]{\distmean}_{\cusvector{\inducingpoints}}, \cusmatrix[bm]{\distcov}_{\cusvector{\inducingpoints}}$) are then optimized by maximizing the Evidence Lower Bound (ELBO) as
\begin{equation}\label{eq:svgp_elbo}
    \mathcal{L}_{\substack{\text{SVGP} \\ \text{ELBO}}} = \sum^{\Numobservation}_{\numobservation=1} \mathbb{E}_{\variationalprob(\processfunction^{(\objnode)}_{\numobservation})}[\log \probability(\obsouts^{(\objnode)}_{\numobservation}\vert\processfunction^{(\objnode)}_{\numobservation})]-\klconst \text{KL}\big(\variationalprob(\cusvector{\inducingpoints})\Vert\probability(\cusvector{\inducingpoints})\big).
\end{equation}
\subsection{Gaussian Process Networks (GPNs)}\label{sec:gpn}
To model the structured processes, Gaussian Process Network (GPN)~\cite{astudillo2021bayesian} represents the process $\Process$ as a Directed Acyclic Graph (DAG) $\graph$. Each node $\numprocess\in\numprocessset$ in the graph corresponds to a subprocess $\process^{(\numprocess)}$ and is modeled as an independent GP. A key assumption in GPNs is that the noisy observations $\cusvector{\obsouts}^{\nodeparent{\numprocess}}$ from parent nodes are used as direct inputs to their child nodes, thus modeling the process as a network of GPs, $\Mgpnetwork$. Each node's GP is trained on its respective input-output pairs, $\left(\cusvector{\inpvars}^{(\numprocess)}, \cusvector{\obsouts}^{\nodeparent{\numprocess}}\right)$ and $\cusvector{\obsouts}^{(\numprocess)}$, by maximizing the MLL as in~\eqref{eq:STGP_MLL_exact}. Each node can be represented as a tuple
\begin{equation}\label{eq:gpn_subprocess_tuple}
    \process^{(\numprocess)} = \left\langle
    \probability\big(\cusvector{\processfunction}^
    {(\numprocess)}\vert\cusvector{\inpvars}^
    {(\numprocess)}, \cusvector{\obsouts}^{\nodeparent
        {\numprocess}}\big),\;
    \probability(\cusvector{\obsouts}^{(\numprocess)}
    \vert\cusvector{\processfunction}^{(\numprocess)})
    \right\rangle.
\end{equation}

\subsection{Partially Observable Gaussian Process Networks (POGPNs)}\label{sec:pogpn}
The GPN's assumption that noisy parent observations $\cusvector{\obsouts}^{\nodeparent{\numprocess}}$ are the direct inputs to child nodes can be limiting, as it fails to distinguish between the true underlying process output and measurement noise~\cite{kiroriwal2025partially}. The Partially Observable Gaussian Process Network (POGPN), developed by Kiroriwal et al.~\cite{kiroriwal2025partially}, provides a more general and robust framework. As shown in Figure~\ref{fig:multi_process}, it is not the noisy observation $\cusvector{\obsouts}^{\nodeparent{\numprocess}}$ that becomes the input but rather the latent output $\cusvector{\trueouts}^{\nodeparent{\numprocess}}$. Using this as the key idea, POGPN models each subprocess (node) as an SVGP, where each node can be represented as a tuple
\begin{equation}\label{eq:pogpn_subprocess_tuple}
    \process^{(\numprocess)} = \left\langle
    \probability\big(\cusvector{\processfunction}^
    {(\numprocess)}\vert\cusvector{\inpvars}^
    {(\numprocess)}, \cusvector{\trueouts}^{\nodeparent
        {\numprocess}}\big),\;
    \probability(\cusvector{\obsouts}^{(\numprocess)}
    \vert\cusvector{\processfunction}^{(\numprocess)})
    \right\rangle.
\end{equation}
Due to POGPN's hierarchical structure, exact inference is intractable. Instead, POGPN uses doubly stochastic variational inference (DSVI)~\cite{salimbeni2017doubly} to optimize the entire network jointly. Kiroriwal et al.~\cite{kiroriwal2025partially} formulate the ELBO, $\mathcal{L}_{\substack{\text{POGPN} \\\text{ELBO}}}$, as
% \begin{align}\label{eq:pogpn_elbo}
%     \mathcal{L}_{\substack{\text{POGPN}                                                                                                                                                                                            \\\text{ELBO}}}^{(\numprocessset)} = &\underbrace{\sum_{\numprocess\in\numprocessset}\sum^{\Numobservation^{(\numprocess)}}_{\numobservation=1}\mathbb{E}_{\variationalprob(\cusvector{\processfunction}^{(\numprocess)}_{\numobservation})}\big[\log\probability(\cusvector{\obsouts}^{(\numprocess)}_{\numobservation}\vert\cusvector{\processfunction}^{(\numprocess)}_{\numobservation})\big]}_{\text{LL loss}}\nonumber \\
%      & -  \underbrace{\klconst\sum_{\numprocess\in\numprocessset}\text{KL}\big(\variationalprob(\cusvector{\inducingpoints}^{(\numprocess)})\Vert\probability(\cusvector{\inducingpoints}^{(\numprocess)})\big)}_{\text{KL loss}},
% \end{align}
\begin{equation}\label{eq:pogpn_elbo}
    \underbrace{\sum_{\numprocess\in\numprocessset}\sum^{\Numobservation^{(\numprocess)}}_{\numobservation=1}\mathbb{E}_{\variationalprob(\cusvector{\processfunction}^{(\numprocess)}_{\numobservation})}\big[\log\probability(\cusvector{\obsouts}^{(\numprocess)}_{\numobservation}\vert\cusvector{\processfunction}^{(\numprocess)}_{\numobservation})\big]}_{\text{LL loss}}-  \underbrace{\klconst\sum_{\numprocess\in\numprocessset}\text{KL}\big(\variationalprob(\cusvector{\inducingpoints}^{(\numprocess)})\Vert\probability(\cusvector{\inducingpoints}^{(\numprocess)})\big)}_{\text{KL loss}},
\end{equation}
where $\cusvector{\inducingpoints}^{(\numprocess)}$ are the inducing points for the SVGP at each node. Inspired by~\cite{jankowiak2020parametric},~\cite{kiroriwal2025partially} also propose a Predictive Log Likelihood (PLL) for POGPN where the `KL loss' remains the same as in~\eqref{eq:pogpn_elbo} and the `LL loss' term is defined as
\begin{equation}\label{eq:pogpn_pll}
    \text{LL}_{\text{PLL}} = \underbrace{\sum_{\numprocess\in\numprocessset}\sum^{\Numobservation^{(\numprocess)}}_{\numobservation=1}\log\mathbb{E}_{\variationalprob(\cusvector{\processfunction}^{(\numprocess)}_{\numobservation})}\big[\probability(\cusvector{\obsouts}^{(\numprocess)}_{\numobservation}\vert\cusvector{\processfunction}^{(\numprocess)}_{\numobservation})\big]}_{\text{Predictive Log-Likelihood}}.
\end{equation}

\subsection{Bayesian Optimization: Expected Improvement (EI)}\label{sec:ei_log}
In Bayesian optimization, Expected Improvement (EI) is a popular acquisition function. Given the best-observed output, $\obsouts^{(\objnode)}_{\incumbentobs}$, the EI at a candidate point $\cusvector{\inpvars}_{\boiter}$ is the expected value of the improvement over the best observed output and is defined as $\text{EI}(\cusvector{\inpvars}_{\boiter}) = \mathbb{E}_{\probability(\obsouts\vert\cusvector{\inpvars}_{\boiter},\mathcal{D})}\left[\max\left(0, \obsouts^{(\objnode)}_{\boiter} - \obsouts^{(\objnode)}_{\incumbentobs}\right)\right].$ When the model's predictive posterior is Gaussian, $\mathcal{N}(\distmean_{\probability(\processfunction^{(\objnode)}_{\boiter})}, \distcov_{\probability(\processfunction^{(\objnode)}_{\boiter})})$, EI has a closed-form solution~\cite{frazier2018tutorial} as
\begin{equation}\label{eq:EI_closed_form}
    \mathrm{EI}(\cusvector{\inpvars}) = \distcov_{\probability(\processfunction^{(\objnode)}_{\boiter})}\,\heifunc\!\left(\frac{\distmean_{\probability(\processfunction^{(\objnode)}_{\boiter})} - \obsouts^{(\objnode)}_{\incumbentobs}}{\distcov_{\probability(\processfunction^{(\objnode)}_{\boiter})}}\right),
    \;\; \heifunc(\eiconst)=\phi(\eiconst)+\eiconst\,\Phi(\eiconst),
\end{equation}
where $\eiconst = (\distmean_{\probability(\processfunction_{\boiter})} - \obsouts_{\incumbentobs})/\distcov_{\probability(\processfunction_{\boiter})}$, and $\Phi(\cdot)$ and $\phi(\cdot)$ are the Cumulative Distribution Function (CDF) and Probability Density Function (PDF) of the standard normal distribution, respectively. Numerical issues of EI were first analyzed by~\cite{ament2023unexpected}, and they proposed Log Expected Improvement (LogEI) as $\text{LogEI}(\cusvector{\inpvars}) = \log \heifunc(\eiconst) + \log \distcov_{\probability(\processfunction^{(\objnode)}_{\boiter})}$ as a robust and stable alternative to EI.

\subsection{Inducing Point Allocator (IPA)}\label{sec:inducing_points_allocator}
The placement of inducing locations significantly impacts the performance of an SVGP~\cite{burt2020convergence}. Initialization strategy Greedy Variance Reduction (GVR), introduced by~\cite{burt2020convergence}, iteratively selects points from the observed inputs $\cusmatrix{\Inpvars}$ that maximally reduce the model's variance: $\cusvector{\inducinglocations}_{\numinducing} = \arg \max_{\cusvector{\inducinglocations} \in \cusmatrix{\Inpvars}} \gvrsymbol_{\numinducing-1}(\cusvector{\inducinglocations})$, where $\gvrsymbol^{2}_{\numinducing-1}(\cusvector{\inducinglocations}) = \gpkernel(\cusvector{\inducinglocations}, \cusvector{\inducinglocations}) -\gpkernel(\cusmatrix{\Inducinglocations}_{1:\numinducing-1},\cusvector{\inducinglocations})^{T} \gpkernel(\cusmatrix{\Inducinglocations}_{1:\numinducing-1}, \cusmatrix{\Inducinglocations}_{1:\numinducing-1})^{-1}\gpkernel(\cusmatrix{\Inducinglocations}_{1:\numinducing-1}, \cusvector{\inducinglocations})$.

Moss et al.~\cite{moss2023inducing} showed that GVR is not optimal for Bayesian optimization using SVGP and introduced a quality function $\girsymbol(\cdot)$ into the selection criterion $\cusvector{\inducinglocations}_{\numinducing} = \arg \max_{\cusvector{\inducinglocations} \in \cusmatrix{\Inpvars}} \girsymbol(\cusvector{\inducinglocations})\gvrsymbol_{\numinducing-1}(\cusvector{\inducinglocations})$. For improved BO performance, they proposed Greedy Improvement Reduction (GIR) that uses an EI-inspired quality function $\girsymbol(\cusvector{\inpvars}_{\numobservation})=\obsouts^{(\objnode)}_{\numobservation}-\obsouts^{(\objnode)}_{\incumbentobs}$.
\section{Related Work}\label{sec:related_work}
Bayesian optimization (BO) with a Gaussian Process (GP)~\cite{rasmussen2003gaussian} surrogate model is an effective methodology for manufacturing processes, with applications ranging from chemical synthesis~\cite{Shields2021Bayesian} to mechanical engineering~\cite{Chung2022A, Brasington2022Bayesian}. Research in this direction can be broadly categorized into three main categories: (1) development of robust acquisition functions, (2) hierarchical surrogate models using process structure and intermediate data, and (3) improving the scalability of standard GPs to handle high-dimensional inputs.

Works by~\cite{Shields2021Bayesian, Brasington2022Bayesian, liang2021scalable} belong to the
first category, in which the authors develop robust acquisition functions either using trust
region or other numerically stable techniques to apply existing BO methods for optimizing
complex processes. In particular, TuRBO~\cite{liang2021scalable} and LogEI~\cite
{ament2023unexpected} acquisition functions have emerged as a consistent, robust choice
across diverse BO applications.

In the second category, motivated by the fact that many manufacturing processes consist of interconnected subprocesses, Gaussian Process Networks (GPNs)~\cite{astudillo2021bayesian,aglietti2020causal,sussex2022model} model the system as a DAG, using intermediate observations from each subprocess to improve the surrogate's accuracy and, consequently, optimization performance. For instance, causal BO with GPNs was applied to optimize fertilizer dosage in farming with respect to the trade-off between crop growth and soil health~\cite{aglietti2020causal}. However, as shown by~\cite{kiroriwal2025partially}, GPNs assume that the GPs at each node can be trained conditionally independently, resulting in suboptimal inference. To address this, the Partially Observable GPN (POGPN) uses variational inference to learn a latent distribution of the process network~\cite{kiroriwal2025partially}. Results show that POGPN can improve BO performance by incorporating partial observations during both training and inference. However, both GPNs and POGPNs are limited in their ability to handle high-dimensional data, such as sensor time series.

A third direction of research by~\cite{hvarfner2024vanilla} aims to enhance standard single-task GP for high-dimensional inputs by using dimensionally-scaled priors for the GP kernel and likelihood. Combined with the robust LogEI~\cite{ament2023unexpected} acquisition function, the results from~\cite{hvarfner2024vanilla} provide robust results for not only high-dimensional input processes. While the method is effective, it relies on a direct input-output mapping and it does not incorporate the high-dimensional intermediate observations $\cusmatrix{\Obsstates}^{(\numprocess)}$ from subprocesses.

Joint Parameter and State-Space (JPSS) modeling was introduced to handle high-dimensional state-space observations $\left(\cusmatrix{\Obsstates}^{(\numprocess)}\right)_{\numprocess\in\numprocessset}$ by using expert knowledge to extract low-dimensional features $\cusvector{\features}=\left(\cusvector{\features}^{(\numprocess)}\right)_{\numprocess\in\numprocessset}$ for a GPN~\cite{kiroriwal2024joint}. However, the original JPSS framework was proposed using a standard GPN, thus inheriting its limitations in uncertainty propagation. This paper integrates POGPN inference with JPSS to optimize process networks, such as the multi-stage bioethanol production process, with high-dimensional intermediate observations.
\section{Motivating Example}\label{sec:problem_description}
The bioethanol production process, illustrated in Fig.~\ref{fig:multi_process}, uses a seed train fermentation technology~\cite{hernandez2013seed}. This is done by scaling up a microbial culture in a small laboratory flask to a large scale in three stages \(\{ \process^{(\numprocess) } \}_{\numprocess \in \{1, 2, 3 \}}\). This scaling up is called a seed train because of its sequential nature. This gives rise to a tightly coupled, multi-stage dynamical system in which the performance of the early stages critically determines the final outcome.
\begin{figure}
    \centering
    \adjustbox{width=0.995\columnwidth}{
        \input{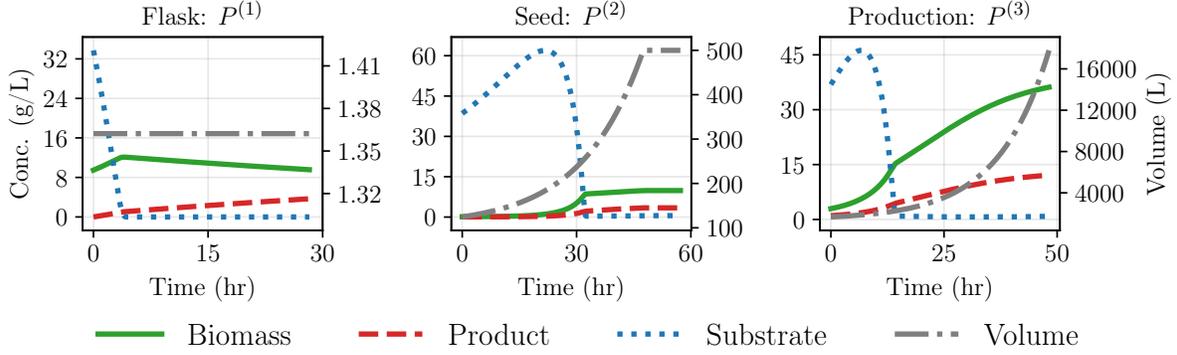}
    }
    \caption{Observed state-space dynamics from a single simulation run of the multi-stage bioethanol production process.}
    \label{fig:bioethanol_states}
\end{figure}

The whole process, denoted by $\Process$, is governed by an input vector with a dimensionality of $\dims{\cusvector{\inpvars}}=26$, denoted by $\cusvector{\inpvars}$, with dimensionalities $\dims{\cusvector{\inpvars}^{(1)}} = 6, \dims{\cusvector{\inpvars}^{(2)}} = 10,$ and $\dims{\cusvector{\inpvars}^{(3)}} = 10$, respectively. The input variables represent operating conditions, including substrate feed rate profiles, linear temperature and pH profiles, and stage durations. The objective function is to optimize the controllable input variables $\cusvector{\inpvars}$ to achieve maximum final process productivity $\obsouts^{(\objnode)}$, quantified as Space-Time Yield (STY)~\cite{hernandez2013seed}. STY, defined as the amount of ethanol produced per unit of time and reactor volume, penalizes inefficiently long stages and thus drives the optimization toward overall process efficiency.

The sequence begins with \emph{$\process^{(1)}$ (Shake Flask)}, which involves activating the biomass originally obtained from a lab stock using a 2L flask. The next step is \emph{$\process^{(2)}$ (Seed Fermenter)}, which cultivates the activated biomass in a 500 L bioreactor operating in fed-batch mode to initiate a dense cell culture ideal for production. The objective is to allow healthy cell growth by precisely manipulating factors such as temperature, pH, and substrate addition to avoid inhibitory effects. Then, final ethanol productions take place during \emph{$\process^{(3)}$ (Production Fermenter)}, a 20000L bioreactor, which utilizes the biomass obtained from culturing to break down a substrate on a larger scale to produce ethanol.

\emph{Process Dynamics and Control:} The dynamics of each subprocess are governed by a set of ordinary differential equations (ODEs) that describe the evolution of concentration of biomass ($\obsstates_\text{biomass}$), ethanol ($\obsstates_\text{ethanol}$),
and  substrate ($\obsstates_\text{substrate}$), and the culture volume ($\obsstates_\text{volume}$)~\cite{salihu2022effect} as
\begin{align}
    \label{eq:bioethanol_dynamics}
    \frac{d\obsstates_\text{biomass}}{dt}   & = (\mu_\text{growth} - k_d)\obsstates_\text{biomass} - \frac{F_{\text{in}}}{\obsstates_\text{volume}} \obsstates_\text{biomass} \nonumber                        \\
    \frac{d\obsstates_\text{ethanol}}{dt}   & = q_{\text{ethanol}} \obsstates_\text{biomass} - \frac{F_{\text{in}}}{\obsstates_\text{volume}} \obsstates_\text{ethanol} \nonumber                              \\
    \frac{d\obsstates_\text{substrate}}{dt} & = \frac{F_{\text{in}}}{\obsstates_\text{volume}} (\obsstates_\text{in} - \obsstates_\text{substrate}) - q_{\text{substrate}} \obsstates_\text{biomass} \nonumber \\
    \frac{d\obsstates_\text{volume}}{dt}    & = F_{\text{in}},
\end{align}
where $k_d$ is the death rate, $F_{\text{in}}$ is the feed rate, $\obsstates_\text{in}$ is the inlet substrate concentration, $\obsstates_\text{volume}$ is volume of the complete culture in the reactor, and $\mu_\text{growth}$ is the specific biomass growth rate. The specific ethanol production rate $q_{\text{ethanol}}$ is described by Luedeking-Piret model, and the specific substrate consumption rate $q_{\text{substrate}}$ follows the Pirt model, which accounts for substrate used for growth, production, and cell maintenance~\cite{luedeking1959kinetic, pirt1965maintenance}.

\emph{The Optimization Challenge:} The bioethanol seed train $\Process$ is a challenge to optimize because of several factors, such as:

\begin{itemize}
    \item \emph{Complex Biological Kinetics:} The dynamics are non-linear and involve growth- and non-growth-related product formation (Luedeking-Piret kinetics) \cite{luedeking1959kinetic}, substrate consumption related to cell maintenance (Pirt model)~\cite{pirt1965maintenance}, and strong inhibitory effects due to high substrate concentration (Andrews model)~\cite{andrews1968} and inhibitory effects due to high ethanol concentration (Ghose-Tyagi model)~\cite{ghose1979rapid}.

    \item \emph{Critical Trade-offs Across Stages:} The optimizer has to make trade-offs between multiple objectives. For instance, minimizing the seed stages reduces overall process time but can result in low-quality inoculum, which negatively affects final productivity~\cite{hernandez2013seed}. A very aggressive feeding mode to enhance productivity can lead to substrate and product inhibitions, causing retardation in the final process.

    \item \emph{Tightly Coupled Dynamics:} The decisions that are made within the early stages impact other stages. This complicates the problem significantly because it is non-trivial to optimize variables within a control system at the holistic level.
\end{itemize}

These factors make the bioethanol seed train a highly suitable and demanding benchmark for assessing structure-aware optimization algorithms. An example of the observed state-space from different stages of the bioethanol production process is shown in Figure~\ref{fig:bioethanol_states}. A threshold STY required to make this pilot-scale simulation economically viable can be 0.5. Importantly, this state-space representation provides information on how long each of these processes is supposed to take. Notably, the optimizer must also determine the optimal duration for each subprocess—$\process^{(1)}$, $\process^{(2)}$, and $\process^{(3)}$—which can range from 6-48 hours, 12-96 hours, and 48-168 hours, respectively. Given that each simulation represents a very time-intensive real-world process, fast optimization of such a time-intensive process is of utmost importance, as it can save a lot of resources and time and hence opportunity costs as well.\footnote{The code for the simulation is available at \url{https://github.com/Sam4896/seed_train_bioethanol_sim}}
\begin{figure*}[t]
    \centering
    \adjustbox{width=0.99\textwidth}{
        \input{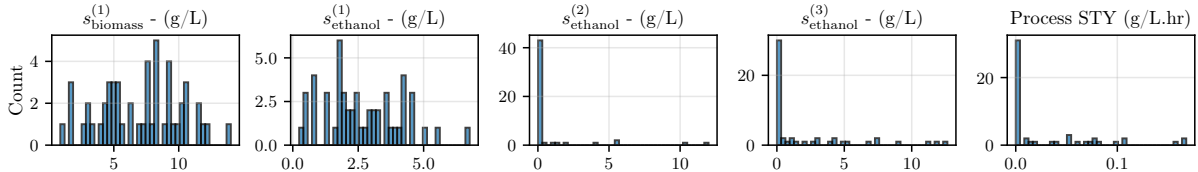}
    }
    \caption{Histograms of key process observations from 50 random simulation runs. The observations shown are the final biomass $\obsstates_\text{biomass}^{(1)}$, ethanol $\obsstates_\text{ethanol}^{(1)}$ concentrations from $\process^{(1)}$, the final ethanol concentrations $\obsstates_\text{ethanol}^{(2)}$ and $\obsstates_\text{ethanol}^{(3)}$ from $\process^{(2)}$ and $\process^{(3)}$, and the final Space-Time Yield (STY) $\obsouts^{(\objnode)}$ for the entire process.}
    \label{fig:bioethanol_process_observations}
\end{figure*}
% \clearpage
\section{Methodology}\label{sec:results}
As explained in Section~\ref{sec:problem_description}, optimizing the bioethanol production process is a significant challenge. In this section, we evaluate the performance of our proposed POGPN-JPSS approach, which integrates Partially Observable Gaussian Process Networks (POGPN)~\cite{kiroriwal2025partially} with Joint Parameter and State Space (JPSS) representation~\cite{kiroriwal2024joint}. We compare our method against three state-of-the-art models: a standard GPN~\cite{astudillo2021bayesian}, a high-dimensional single-task GP (STGP)~\cite{hvarfner2024vanilla}, and a standard SVGP~\cite{moss2023inducing}. All optimizations are driven by the robust Log Expected Improvement (LogEI) acquisition function~\cite{ament2023unexpected}.

\begin{figure*}
    \centering
    \begin{subfigure}[c]{0.38\textwidth}
        \centering
        \adjustbox{width=0.99\textwidth}{
            \tikzset{every picture/.style={line width=0.7pt}} % Set default line width
\begin{tikzpicture}[node distance=10mm and 15mm,
        process/.style={circle, draw, minimum size=7mm, align=center, inner sep=0.5pt},
        every node/.style={font=\footnotesize},
        arr/.style={-Stealth},
    ]
    % Define nodes
    \node[process] (p1) {$\cusvector{\features}^{(1)}$};
    \node[process] (p2) [right=of p1, xshift=-0.5cm] {$\features^{(2)}$};
    \node[process] (p3) [right=of p2, xshift=-0.5cm] {$\features^{(3)}$};
    \node[process] (obj) [right=of p3, xshift=-0.5cm] {$\obsouts^{(\objnode)}$};

    \node[process] (x1) [below=of p1, yshift=0.6cm] {$\cusvector{\inpvars}^{(1)}$};
    \node[process] (x2) [below=of p2, yshift=0.6cm] {$\cusvector{\inpvars}^{(2)}$};
    \node[process] (x3) [below=of p3, xshift=9mm, yshift=0.6cm] {$\cusvector{\inpvars}^{(3)}$};

    % Arrows
    \draw[arr] (x1) -- (p1);
    \draw[arr] (p1) -- (p2);
    \draw[arr] (x2) -- (p2);
    \draw[arr] (p2) -- (p3);
    \draw[arr] (p3) -- (obj);
    \draw[arr] (p2.north) to[bend left=15] (obj.north);
    \draw[arr] (x3.north) to[out=160, in=-90] (p3.south);
    \draw[arr] (x3.north) to[out=20, in=-90] (obj.south);

\end{tikzpicture}
        }
        \caption*{DAG for the bioethanol production process.}
        \label{fig:bioethanol_dag}
    \end{subfigure}%
    \hfill
    \begin{subfigure}[c]{0.6\textwidth}
        \centering
        \adjustbox{width=0.99\textwidth}{
            \input{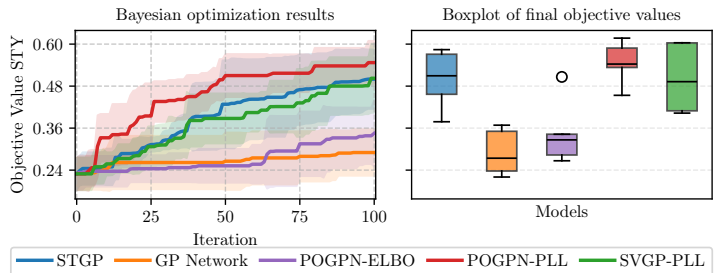}
        }
        % \caption{Optimization results for the bioethanol production process.}
        \label{fig:bioethanol_optimization_results}
    \end{subfigure}
    \caption{The Directed Acyclic Graph (DAG) for the bioethanol process is shown alongside the Bayesian optimization performance of different surrogate models. The plots show the mean and standard deviation of the best-observed Space-Time Yield (STY) with an optimization budget of 100 iterations.}
    \label{fig:bioethanol_dag_and_results}
\end{figure*}

The state-space observations $\cusmatrix{\Obsstates}^{(1)}$, $\cusmatrix{\Obsstates}^{(2)}$, and $\cusmatrix{\Obsstates}^{(3)}$ from each stage are high-dimensional multivariate time-series. We reduce these to low-dimensional features using the JPSS approach, incorporating process expert knowledge to identify features essential to the next stages and the final objective. The final objective in this case is to optimize STY, which depends on total ethanol produced, process volume, and total duration. For $\process^{(1)}$, our objective is to optimize a healthy inoculum to be introduced into the seed fermenter. Both biomass and ethanol concentrations are key indicators of culture health, as ethanol can inhibit growth. Therefore, we extract the final biomass and ethanol concentrations, $\cusvector{\features}^{(1)}=\left(\obsstates_\text{biomass}^{(1)},\obsstates_\text{ethanol}^{(1)}\right)$, as the intermediate features. For $\process^{(2)}$ and $\process^{(3)}$  our objective is to optimize ethanol production, so we use the final ethanol concentrations, $\features^{(2)}=\obsstates_\text{ethanol}^{(2)}$ and $\features^{(3)}=\obsstates_\text{ethanol}^{(3)}$, as the respective intermediate observations. STY of the entire process depends on the total ethanol output, process volume, and total duration. The DAG used for our structured models, as shown in Figure~\ref{fig:bioethanol_dag_and_results}.

To better understand this optimization challenge, we show the histogram plots of the discussed intermediate observations for 50 randomly performed simulation runs in Figure~\ref{fig:bioethanol_process_observations}. Although the observations $\cusvector{\features}^{(1)}$ are relatively spread out, outputs observed from the other subprocess and the STY are mainly concentrated close to zero and have a long tail. This kind of distribution is particularly suggestive that regions corresponding to high-performing STY are relatively sparse within the 26-dimensional input space.

\section{Benchmark Results}\label{sec:benchmark_results}
We compare the performance of POGPN-JPSS (using both ELBO and PLL objectives) agains state-of-the-art models: GPN~\cite{astudillo2021bayesian}, STGP~\cite{hvarfner2024vanilla}, and SVGP-PLL~\cite{moss2023inducing} with PLL objective~\cite{jankowiak2020parametric}. All models were implemented in BoTorch~\cite{balandat2020botorch} with a Matern-5/2 kernel, a Gaussian likelihood, and the dimensionally-scaled priors proposed by~\cite{hvarfner2024vanilla}. For POGPN, the number of inducing points per node was set to the number of training data points for that node. Inducing points for the objective node were initialized using Greedy Improvement Reduction (GIR)~\cite{moss2023inducing}, while all other nodes used Greedy Variance Reduction (GVR)~\cite{burt2020convergence}. We used the Adam optimizer~\cite{kingma2014adam} with a learning rate of 0.02 and an exponential moving average stopping criterion. All observations were log-transformed and standardized before model training to improve numerical stability and the exploration-exploitation trade-off. Each optimization had a budget of 100 iterations, starting with an initial dataset of 50 randomly sampled points, and all experiments were repeated 5 times to report the average performance.

As depicted in Figure~\ref{fig:bioethanol_dag_and_results}, POGPN-PLL is able to achieve an optimal solution for bioethanol production much faster and more robustly than any other state-of-the-art methods. POGPN-PLL provides a higher STY value with reduced variability. Most importantly, POGPN-PLL can reach the 0.5 STY threshold in 50 iterations, whereas other best approaches require almost 100 iterations. Given the duration of the subprocess, as discussed in Section~\ref{sec:problem_description}, this accelerated optimization results in an average savings of 300-350 days of production time, representing a substantial reduction in opportunity costs, materials, and energy use, while increasing overall productivity.
\section{Discussion and Outlook}\label{sec:discussion}
We show that the Partially Observable Gaussian Process Network - Predictive Log-Likelihood (POGPN-PLL), when combined with the Joint Parameter and State-Space (JPSS), significantly outperforms state-of-the-art models in optimizing the complex, multi-stage bioethanol production process. Faster, more robust convergence to the optimal solution can be attributed to POGPN-PLL's ability to learn the problem's latent structure. It is achieved by creating a process network by utilizing process expert knowledge to extract low dimensional features from the high-dimensional state-space. The rapid optimization translates directly into economic benefits, including reduced opportunity costs and savings in materials and energy.

Process expert knowledge in the form of feature extraction enables the model to exploit high-dimensional sensor data and the process structure, leading to faster, more robust convergence to the optimal solution. POGPN is a promising framework for Bayesian optimization of real-world industrial processes, where time and material efficiency are essential.

Future work could investigate the performance gap between the PLL and the standard Evidence Lower Bound (ELBO) objective for POGPN, which may be attributed to how each objective learns the process stochasticity. It would also be interesting to explore feature-extraction and causal-graph-discovery methods combined with POGPN-JPSS for Bayesian optimization, thereby extending POGPN's applicability to systems where the structure is not well-understood.

The research in this paper was supported by the Deutsche Forschungsgemeinschaft (DFG, German Research Foundation) - 459291153 (FOR5399).

\bibliography{setup/references}

@article{astudillo2021bayesian,
  title={Bayesian optimization of function networks},
  author={Astudillo, Raul and Frazier, Peter},
  journal={Advances in neural information processing systems},
  volume={34},
  pages={14463--14475},
  year={2021}
}

@inproceedings{aglietti2020causal,
  title={Causal bayesian optimization},
  author={Aglietti, Virginia and Lu, Xiaoyu and Paleyes, Andrei and Gonz{\'a}lez, Javier},
  booktitle={International Conference on Artificial Intelligence and Statistics},
  pages={3155--3164},
  year={2020},
  organization={PMLR}
}

@book{goldratt2006goal,
  title={The goal},
  author={Goldratt, Eliyahu M and Cox, Jeff and Whitford, David},
  year={2006},
  publisher={HighBridge}
}

@article{kingma2014adam,
  title={Adam: A method for stochastic optimization},
  author={Kingma, Diederik P and Ba, Jimmy},
  journal={arXiv preprint arXiv:1412.6980},
  year={2014}
}

@article{hvarfner2024vanilla,
  title={Vanilla Bayesian optimization performs great in high dimensions},
  author={Hvarfner, Carl and Hellsten, Erik Orm and Nardi, Luigi},
  journal={arXiv preprint arXiv:2402.02229},
  year={2024}
}

@inproceedings{moss2023inducing,
  title={Inducing point allocation for sparse Gaussian processes in high-throughput Bayesian optimisation},
  author={Moss, Henry B and Ober, Sebastian W and Picheny, Victor},
  booktitle={International Conference on Artificial Intelligence and Statistics},
  pages={5213--5230},
  year={2023},
  organization={PMLR}
}

@article{balandat2020botorch,
  title={BoTorch: A framework for efficient Monte-Carlo Bayesian optimization},
  author={Balandat, Maximilian and Karrer, Brian and Jiang, Daniel and Daulton, Samuel and Letham, Ben and Wilson, Andrew G and Bakshy, Eytan},
  journal={Advances in neural information processing systems},
  volume={33},
  pages={21524--21538},
  year={2020}
}

@article{sano2020application,
  title={Application of Bayesian optimization for pharmaceutical product development},
  author={Sano, Syusuke and Kadowaki, Tadashi and Tsuda, Koji and Kimura, Susumu},
  journal={Journal of Pharmaceutical Innovation},
  volume={15},
  number={3},
  pages={333--343},
  year={2020},
  publisher={Springer}
}

@techreport{ids2025realcost,
  title={The Real Cost of Manufacturing Downtime (2020--2025): Sector Insights \& Forecast},
  author={IDS-INDATA},
  year={2025},
  url={https://idsindata.co.uk/manufacturing-downtime-costs-and-forecasting/},
  note={Accessed: 2025-08-26}
}

@article{sussex2022model,
  title={Model-based causal Bayesian optimization},
  author={Sussex, Scott and Makarova, Anastasiia and Krause, Andreas},
  journal={arXiv preprint arXiv:2211.10257},
  year={2022}
}

@article{salimbeni2017doubly,
  title={Doubly stochastic variational inference for deep Gaussian processes},
  author={Salimbeni, Hugh and Deisenroth, Marc},
  journal={Advances in neural information processing systems},
  volume={30},
  year={2017}
}

@article{hensman2013gaussian,
  title={Gaussian processes for big data},
  author={Hensman, James and Fusi, Nicolo and Lawrence, Neil D},
  journal={arXiv preprint arXiv:1309.6835},
  year={2013}
}

@inproceedings{jankowiak2020parametric,
  title={Parametric gaussian process regressors},
  author={Jankowiak, Martin and Pleiss, Geoff and Gardner, Jacob},
  booktitle={International conference on machine learning},
  pages={4702--4712},
  year={2020},
  organization={PMLR}
}

@incollection{rasmussen2003gaussian,
  title={Gaussian processes in machine learning},
  author={Rasmussen, Carl Edward},
  booktitle={Summer school on machine learning},
  pages={63--71},
  year={2003},
  publisher={Springer}
}

@article{frazier2018tutorial,
  title={A tutorial on Bayesian optimization},
  author={Frazier, Peter I},
  journal={arXiv preprint arXiv:1807.02811},
  year={2018}
}

@article{burt2020convergence,
  title={Convergence of sparse variational inference in Gaussian processes regression},
  author={Burt, David R and Rasmussen, Carl Edward and Van Der Wilk, Mark},
  journal={Journal of Machine Learning Research},
  volume={21},
  number={131},
  pages={1--63},
  year={2020}
}

@article{ament2023unexpected,
  title={Unexpected improvements to expected improvement for bayesian optimization},
  author={Ament, Sebastian and Daulton, Samuel and Eriksson, David and Balandat, Maximilian and Bakshy, Eytan},
  journal={Advances in Neural Information Processing Systems},
  volume={36},
  pages={20577--20612},
  year={2023}
}

@article{andrews1968,
  author    = {J. F. Andrews},
  title     = {A Mathematical Model for the Continuous Culture of Microorganisms Utilizing Inhibitory Substrates},
  journal   = {Biotechnology and Bioengineering},
  volume    = {10},
  number    = {6},
  pages     = {707--723},
  year      = {1968},
  doi       = {10.1002/bit.260100602}
}

@article{sanchez2008bioethanol,
  author    = {O. J. Sánchez and C. A. Cardona},
  title     = {Trends in Biotechnological Production of Fuel Ethanol from Different Feedstocks},
  journal   = {Bioresource Technology},
  volume    = {99},
  number    = {13},
  pages     = {5270--5295},
  year      = {2008},
  doi       = {10.1016/j.biortech.2007.11.013}
}

@article{Shields2021Bayesian,title={Bayesian reaction optimization as a tool for chemical synthesis},author={Benjamin J. Shields and Jason M. Stevens and Jun Li and Marvin Parasram and Farhan N. Damani and Jesus I Martinez Alvarado and J. Janey and Ryan P. Adams and A. Doyle},journal={Nature},year={2021},volume={590},pages={89 - 96},doi={10.1038/s41586-021-03213-y}}

@inproceedings{
liang2021scalable,
title={Scalable Bayesian Optimization Accelerates Process Optimization of Penicillin Production},
author={Qiaohao Liang and Lipeng Lai},
booktitle={NeurIPS 2021 AI for Science Workshop},
year={2021},
}

@article{Brasington2022Bayesian,title={Bayesian optimization for process planning selections in automated fiber placement},author={A. Brasington and J. Halbritter and R. Wehbe and R. Harik},journal={Journal of Composite Materials},year={2022},volume={56},pages={4275 - 4296},doi={10.1177/00219983221129010}}

@article{Chung2022A,title={A Multi-Stage Approach for Knowledge-Guided Predictions With Application to Additive Manufacturing},author={Seokhyun Chung and Cheng-Hao Chou and Xiaozhu Fang and Raed Al Kontar and C. Okwudire},journal={IEEE Transactions on Automation Science and Engineering},year={2022},volume={19},pages={1675-1687},doi={10.1109/tase.2022.3160420}}

@inproceedings{kiroriwal2024joint,
  author={Kiroriwal, Saksham and Pfrommer, Julius and Mende, Hendrik and Schmitt, Robert H. and Beyerer, Jurgen},
  booktitle={2024 IEEE 22nd International Conference on Industrial Informatics (INDIN)}, 
  title={Joint Parameter and State-Space Modelling of Manufacturing Processes using Gaussian Processes}, 
  year={2024},
  volume={},
  number={},
  pages={1-6},
  }

@article{kiroriwal2025partially,
  title={Partially Observable Gaussian Process Network and Doubly Stochastic Variational Inference},
  author={Kiroriwal, Saksham and Pfrommer, Julius and Beyerer, J{\"u}rgen},
  journal={arXiv preprint arXiv:2502.13905},
  year={2025}
}

@article{luedeking1959kinetic,
  title={A kinetic study of the lactic acid fermentation. Batch process at controlled pH},
  author={Luedeking, Robert and Piret, Edgar L},
  journal={Journal of Biochemical and Microbiological Technology and Engineering},
  volume={1},
  number={4},
  pages={393--412},
  year={1959},
  publisher={Wiley}
}

@article{pirt1965maintenance,
  title={The maintenance energy of bacteria in growing cultures},
  author={Pirt, SJ},
  journal={Proceedings of the Royal Society of London. Series B. Biological Sciences},
  volume={163},
  number={991},
  pages={224--231},
  year={1965},
  publisher={The Royal Society London}
}

@article{ghose1979rapid,
  title={Rapid ethanol fermentation of cellulose hydrolysate. II. Product and substrate inhibition and optimization of fermentor design},
  author={Ghose, TK and Tyagi, RD},
  journal={Biotechnology and Bioengineering},
  volume={21},
  number={8},
  pages={1401--1420},
  year={1979},
  publisher={Wiley Online Library}
}

@inproceedings{hernandez2013seed,
  title={Seed train optimization for suspension cell culture},
  author={Hern{\'a}ndez Rodr{\'\i}guez, Tanja and P{\"o}rtner, Ralf and Frahm, Bj{\"o}rn},
  booktitle={BMC Proceedings},
  volume={7},
  pages={P9},
  year={2013},
  organization={Springer}
}

@article{salihu2022effect,
  title={Effect of pH and temperature on bioethanol production: Evidences from the fermentation of sugarcane molasses using Saccharomyces cerevisiae},
  author={Salihu, UY and Usman, UG and Abubakar, AY and Mansir, G},
  journal={Dutse Journal of Pure and Applied Sciences},
  volume={8},
  number={4b},
  pages={9--16},
  year={2022}
}

\end{document}